\documentclass[12pt]{article}

\usepackage[english]{babel}
\usepackage[utf8]{inputenc}
\usepackage[T1]{fontenc}

\usepackage{graphicx}
\usepackage{geometry}
\usepackage{setspace}
\graphicspath{{figs/}}
\usepackage{fancyhdr}

\usepackage{mathtools}
\usepackage{amssymb}
\usepackage{amsmath}
\usepackage{amsthm}
\usepackage{physics}
\usepackage{siunitx}
\usepackage{enumitem}
\theoremstyle{definition}

\usepackage{subcaption}
\usepackage{caption}       
\usepackage{float}
\usepackage{wrapfig}
\usepackage{booktabs}
\usepackage{tikz}
\usepackage{pgfplots}
\pgfplotsset{compat=1.18}
\usepackage{algorithm2e}
\usepackage{listings}

\usepackage{hyperref}
\hypersetup{
    colorlinks=true,
    linkcolor=blue,
    citecolor=red,
    urlcolor=cyan,
    pdftitle={Symmetry-Aware Neural Solvers for Periodic Quantum Eigenproblems},
    pdfauthor={Haaris A. Mian}
}
\usepackage{cleveref}
\numberwithin{equation}{section}

\begin{document}

\begin{titlepage}

\begin{center}
   \includegraphics[scale=0.2]{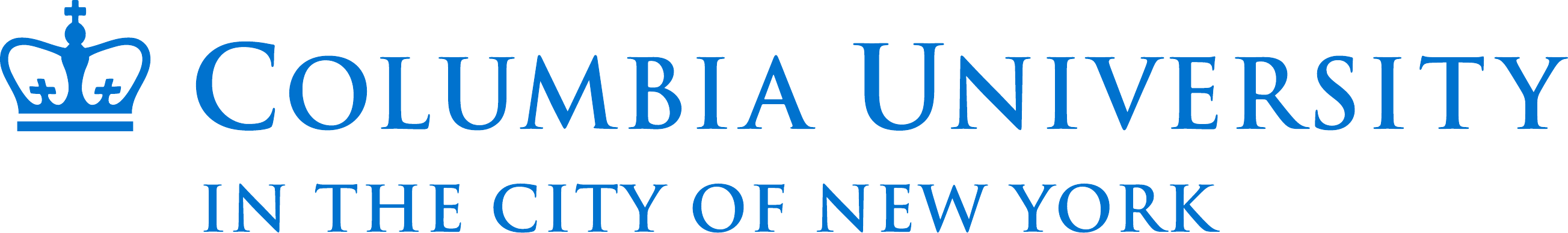} 
\end{center}

\vskip 4cm

\center

\textsc{\large MASTERS THESIS}

\vspace{0.5in}

\noindent\makebox[\linewidth]{\rule{\linewidth}{1.2pt}}
\textsc{ \textbf{\large Physics-Informed Neural Solvers for Periodic Quantum Eigenproblems}} \\
\noindent\makebox[\linewidth]{\rule{\linewidth}{1.2pt}}

\vspace{0.5in}

\begin{minipage}{0.48\textwidth}
    \begin{flushleft}
        \textit{Student:} \\
        Haaris A. Mian\\
    \end{flushleft}
\end{minipage}
\begin{minipage}{0.48\textwidth}
    \begin{flushright}
    \textit{Advisor:} \\
    Professor Michael I. Weinstein \\
    \end{flushright}
\end{minipage}

\vspace{1.5in}

\textbf{\large Program in Applied Mathematics\\[2ex] Department of Applied Physics and Applied Mathematics} \\[4ex]

\today

\end{titlepage}

\newpage
\thispagestyle{empty}
\begin{abstract}
    This thesis presents a physics-informed machine learning framework for solving the Floquet-Bloch eigenvalue problem associated with particles in two-dimensional periodic potentials, with a focus on honeycomb lattice geometry, due to its distinctive band topology featuring Dirac points and its relevance to materials such as graphene. By leveraging neural networks to learn complex Bloch functions and their associated eigenvalues (energies) simultaneously, we develop a mesh-free solver enforcing the governing Schrödinger equation, Bloch periodicity, and normalization constraints through a composite loss function without supervision. The model is trained over the Brillouin zone to recover band structures and Bloch modes, with numerical validation against traditional plane-wave expansion methods. We further explore transfer learning techniques to adapt the solver from nearly-free electron potentials to strongly varying potentials, demonstrating its ability to capture changes in band structure topology. This work contributes to the growing field of physics-informed machine learning for quantum eigenproblems, providing insights into the interplay between symmetry, band structure, and neural architectures.
\end{abstract}
\newpage
\thispagestyle{empty}
\tableofcontents

\newpage
\setcounter{page}{1}

\section{Introduction}
Recently, there has been growing interest in leveraging the techniques of scientific machine learning to solve partial differential equations (PDEs) that arise in physics and engineering. One approach that has shown promise is the use of physics-informed neural networks (PINNs), which allow for the incorporation of governing equations and physical constraints directly into the loss function of a neural network \cite{raissi_physics_2017}. This framework has been successfully applied to a variety of problems, including fluid dynamics, materials science, and quantum mechanics \cite{karniadakis2021physics, cai2021physics, jin2020unsupervised}.

In this thesis, we focus on the application of physics-informed machine learning to the problem of solving the time-independent Schrödinger equation for particles in a periodic potential, an eigenvalue problem of fundamental importance in condensed matter physics and materials science. The theory of electrons in periodic potentials, described by Floquet-Bloch theory, underpins our understanding of the electronic properties of crystalline solids \cite{ashcroft_solid_2011}. Notably, certain lattice structures, such as the honeycomb lattice found in graphene, exhibit conical degeneracies in their band structure known as Dirac points \cite{Castro_Neto_2009, fefferman_honeycomb_2012,Lee_Thorp_2018}. These points give rise to unique electronic properties, including effective massless Dirac fermions, which have been the subject of extensive theoretical and experimental investigation.
\subsection{Contributions and Scope}
This work is motivated by the desire to develop an unsupervised machine learning approach for the Floquet-Bloch eigenvalue problem, with a particular focus on honeycomb lattice potentials and their associated Dirac points. Traditional numerical methods, such as finite element methods, can be computationally expensive for high-dimensional settings and rely on a mesh discretization of the domain. By leveraging the universal approximation capabilities of neural networks, as well as their ability to incorporate physical constraints, we aim to create mesh-free solvers that can recover both eigenvalues and eigenfunctions of the Schrödinger equation with high accuracy. The code accompanying this thesis is publicly available.\footnote{\url{https://github.com/haarisamian/neural-bloch-eigensolver}}

The main contributions of this thesis are as follows:
\begin{itemize}
    \item A physics-informed learning framework for solving the Floquet-Bloch eigenvalue problem in two-dimensional periodic potentials, with a focus on honeycomb lattice geometry. 
    \item Physics-informed objective enforcing Schrödinger equation residuals, boundary conditions, and normalization constraints for physically valid eigenstates, as well as Bloch inductive bias.
    \item Numerical validation of neural network solutions with traditional plane-wave expansion (i.e. spectral) methods, demonstrating their ability to accurately capture band structure topology.
    \item An investigation of curriculum learning and transfer learning techniques to generalize from nearly-free electron potentials to strongly varying and pseudo-atomic potentials.
\end{itemize}
\section{Background and Literature}
\subsection{Floquet-Bloch Theory}
We introduce some relevant results from the theory of periodic quantum media here as background \cite{ReedSimonIV}. If we choose a linearly independent basis
$\{\mathbf{a_1}, \mathbf{a_2}\}$ of $\mathbb{R}^2$ in real space,
we can consider the lattice spanned by these vectors:
\begin{equation}
    \Lambda = \{c_1\mathbf{a_1} + c_2\mathbf{a_2}: c_1, c_2 \in \mathbb{Z}\},
\end{equation}
as well as the reciprocal space (or \textit{dual}) lattice:
\begin{equation}
    \Lambda^* = \{c_1\mathbf{k_1} + c_2\mathbf{k_2}: c_1, c_2 \in \mathbb{Z}\},
\end{equation}
where the vectors $\{\mathbf{k_1}, \mathbf{k_2}\}$ are chosen to satisfy the relation:
\begin{equation}
    \mathbf{k_i}\cdot\mathbf{a_j} = 2\pi\delta_{ij}.
\end{equation}

We restrict our attention to the
two-dimensional time-independent Schrödinger equation for a particle in a periodic potential:
\begin{equation}\label{eq:schrod}
    -\frac{\hbar^2}{2m} \Delta \psi(\mathbf{x}) + V(\mathbf{x})\psi(\mathbf{x}) = E\psi(\mathbf{x}), \quad V(\mathbf{x}+\mathbf{a_i}) = V(\mathbf{x}).
\end{equation}
$V(\mathbf{x})$ is, in general, an arbitrary scalar potential, but we enforce that it has the same periodicity as the idealized real space lattice. For suitable choice of units, we can write $m = \hbar = 1$ without loss of generality. Note that Eq. \eqref{eq:schrod} defines an eigenproblem for complex-valued state wavefunctions $\psi(\mathbf{x})$ and their associated energies (eigenvalues) $E$, which are guaranteed real by the hermiticity of the Hamiltonian $\mathcal{H}$ in the equivalent statement below:
\begin{equation}\label{eq:hammy}
    \mathcal{H}\psi(\mathbf{x}) = E\psi(\mathbf{x}), \quad \mathcal{H} \equiv [-\frac{1}{2} \Delta + V(\mathbf{x})].
\end{equation}

\begin{figure}[t]
    \centering
    \includegraphics[width=0.5\linewidth]{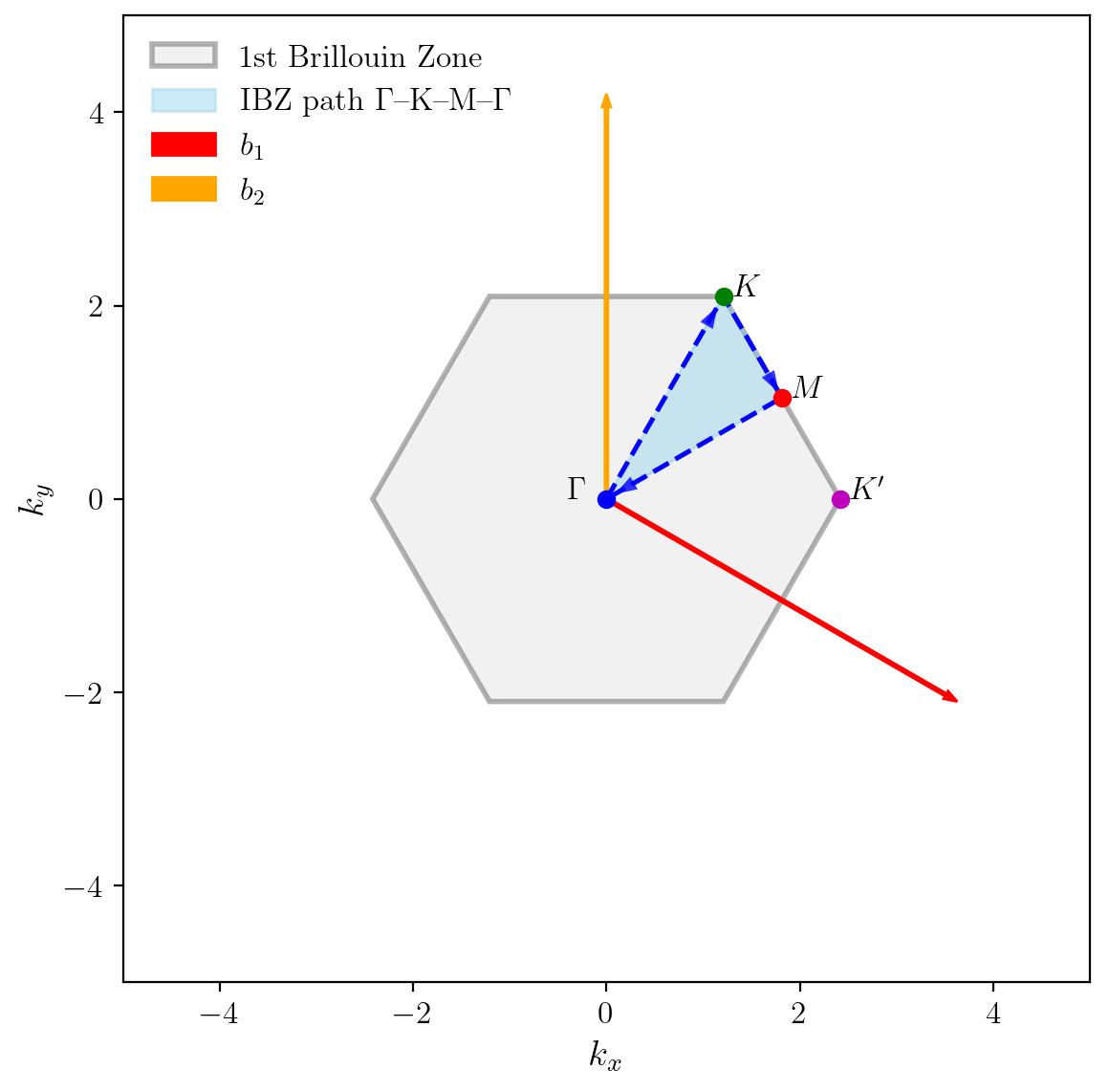}
    \caption{Brillouin zone of the honeycomb lattice and reciprocal lattice vectors \{$b_1$, $b_2$\}. High symmetry points $\Gamma$, K, M, K' labeled, along with the path connecting high-symmetry points over the irreducible wedge of the Brillouin zone.}
    \label{fig:bz}
\end{figure}

Due to the periodicity of the system's potential, Bloch's theorem \cite{ashcroft_solid_2011} allows us to decompose any solution $\psi$ to this problem as the product of a 
plane wave and a $\Lambda$-periodic function, or the \textit{Bloch function} $u_{n,\mathbf{k}}$, as shown below:
\begin{equation}\label{eq:bloch}
    \psi_{n,k}(\mathbf{x}) = e^{i\mathbf{k}\cdot\mathbf{x}}u_{n,\mathbf{k}}(\mathbf{x}).
\end{equation}
In Eq. \eqref{eq:bloch}, $n$ indexes the spectrum of allowable energies, and $\mathbf{k}$ corresponds to the \textit{quasi-momentum} of the particle, which is unique in the reciprocal space up to translation with $\mathbf{k_i}$. Also, due to the $\Lambda$ periodicity of the Bloch function, we obtain the following \textit{pseudo-periodic} boundary condition satisfied by $\psi$: 
\begin{equation}\label{eq:pseudo}
    \psi_{n,k}(\mathbf{x} + \mathbf{a_i}) = e^{i\mathbf{k}\cdot\mathbf{a_i}}\psi(\mathbf{x}).
\end{equation}
Therefore, with these boundary conditions, we are interested in solving the eigenproblem with $\mathbf{x}$ restricted to the real space $\Lambda$ unit cell and $\mathbf{k}$ restricted to the reciprocal space primitive cell known as the \textit{Brillouin zone}, such as in Figure \ref{fig:bz}. In particular, the Brillouin zone encompasses all the points in reciprocal space which are closer to the origin than to any point on the reciprocal latice defined by $\{\mathbf{k_i}\}$.
Now, substituting Eq. \eqref{eq:bloch} into Eq. \eqref{eq:hammy}, we obtain a modified Hamiltonian operator which allows us to solve for the Bloch states directly:
\begin{equation}\label{eq:mod}
    \mathcal{H}_bu_{n,\mathbf{k}}(\mathbf{x}) = E_{n, \mathbf{k}}u_{n,\mathbf{k}}(\mathbf{x}), \quad \mathcal{H}_b \equiv [-\frac{1}{2} (\nabla + i\mathbf{k})^2 + V(\mathbf{x})],
\end{equation}
where every value of $\mathbf{k}$ over the Brillouin zone will produce a discrete eigenspectrum indexed by $n$, and these $E_n(\mathbf{k})$ functions are known as dispersion surfaces or energy bands.

\subsection{Honeycomb Lattice Potentials and Dirac Points}
We begin by considering a triangular lattice $\Lambda_h$ spanned by vectors, with arbitrary lattice spacing $a$:
\begin{equation}
    \mathbf{a_1} = a\begin{pmatrix} \frac{\sqrt{3}}{2}  \\ \frac{1}{2}\end{pmatrix},\quad \mathbf{a_2} = a\begin{pmatrix} \frac{\sqrt{3}}{2}  \\ -\frac{1}{2}  \end{pmatrix},\quad a > 0,
\end{equation}
and its corresponding dual lattice $\Lambda_h^*$:
\begin{equation}
    \mathbf{b_1} = q\begin{pmatrix} \frac{1}{2}  \\ \frac{\sqrt{3}}{2}\end{pmatrix},\quad \mathbf{b_2} = q\begin{pmatrix} \frac{1}{2}  \\ -\frac{\sqrt{3}}{2}  \end{pmatrix},\quad q \equiv \frac{4\pi}{a\sqrt{3}}.
\end{equation}
We now define a $\textit{honeycomb lattice potential}$ \cite{fefferman_honeycomb_2012, Lee_Thorp_2018}, as any real-valued potential $V_h: \mathbb{R}^2 \rightarrow \mathbb{R}$ satisfying the following conditions:
\begin{enumerate}[label=(\Roman*)]
    \item $V_h$ is periodic with respect to $\Lambda_h$, i.e. $V_h(\mathbf{x} + \mathbf{a}) = V_h(\mathbf{x}) \forall x \in \mathbb{R}^2$ and $\mathbf{a} \in \Lambda_h$.
    \item $V_h$ is symmetric with respect to inversion, i.e. $V_h(-\mathbf{x}) = V_h(\mathbf{x})$.
    \item $V_h$ is invariant with respect to the rotation matrix $\mathcal{R}$ where $\mathcal{R}$ denotes the counterclockwise rotation in $\mathbb{R}^2$ by $\frac{2\pi}{3}$.
\end{enumerate}
For this family of potentials, it can been shown that the above symmetries imply the existence of conical singularities, or \textit{Dirac points}, in the Floquet-Bloch dispersion surfaces at the high-symmetry points $K$ and $K'$ lying at the vertices of the Brillouin zone \cite{fefferman_honeycomb_2012,Lee_Thorp_2018}. In other words, the dispersion surfaces of the honeycomb lattice exhibit linear band crossings at these high-symmetry points, and the corresponding eigenvalues form a two-fold degenerate spectral pair. The existence of these points give rise to effective massless Dirac fermions in materials such as graphene \cite{Castro_Neto_2009}, which can be modeled as such a 2D atomic lattice with carbon atoms located at the honeycomb lattice positions. 
\begin{figure}[h]
    \centering
    \includegraphics[width=0.6\textwidth]{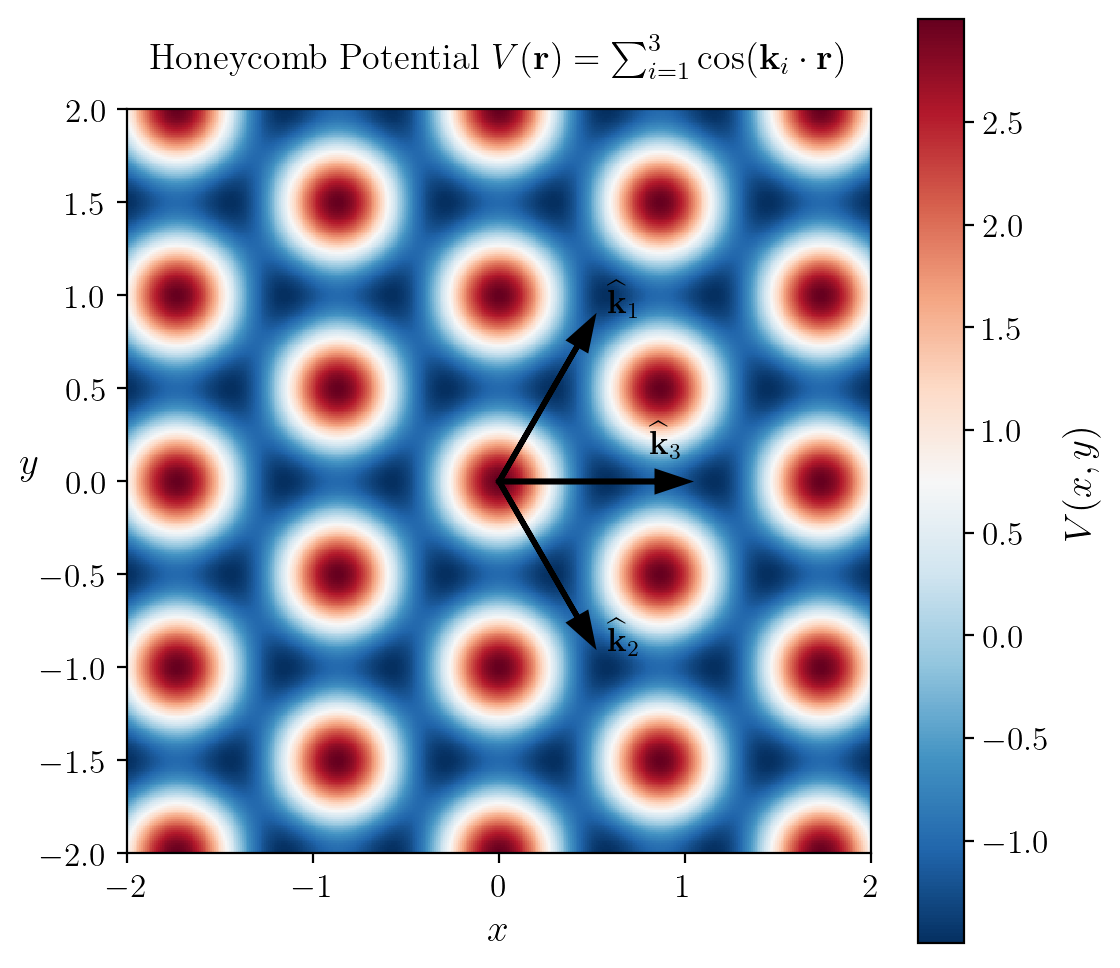}
    \caption{A simple honeycomb lattice potential formed by the superposition of three cosine waves, $V(\mathbf{x}) = V_0 \sum_{i=1}^{3} \cos(\mathbf{b_i}\cdot\mathbf{x})$, where $\mathbf{b_i}$ are the reciprocal lattice vectors.}
    \label{fig:fig1}
\end{figure}
\subsection{Spectral Methods: Numerical Solution by Plane-Waves}
As a ground-truth benchmark, we can employ a convergent numerical solution based on the plane-wave expansion, which yields accurate approximation of eigenvalues and Bloch functions and can be compared to the solutions outputted by the network \cite{kittel2005int,ashcroft_solid_2011}. This approach relies on the Fourier series representation of both the periodic potential and the Bloch functions. For a $\Lambda$-periodic potential $V(\mathbf{x})$, we can write:
\begin{equation}
    V(\mathbf{x}) = \sum_{\mathbf{K} \in \Lambda^*} V_{\mathbf{K}} e^{i\mathbf{K}\cdot\mathbf{x}}, \quad V_{\mathbf{K}} = \frac{1}{|\Omega|}\int_{\Omega} V(\mathbf{x})e^{-i\mathbf{K}\cdot\mathbf{x}} d\mathbf{x},
\end{equation}
where $\Omega$ is the unit cell of the lattice $\Lambda$. Similarly, we can express the Bloch function $u_{n,\mathbf{k}}(\mathbf{x})$ as:
\begin{equation}
    u_{n,\mathbf{k}}(\mathbf{x}) = \sum_{\mathbf{K} \in \Lambda^*} c_{n,\mathbf{k}}(\mathbf{K}) e^{i\mathbf{K}\cdot\mathbf{x}}.
\end{equation}
Substituting these expansions into the modified Hamiltonian eigenproblem in Eq. \eqref{eq:mod} and projecting onto the plane-wave basis leads to a matrix eigenvalue problem for the coefficients $c_{n,\mathbf{k}}(\mathbf{K})$. By truncating the infinite series to a finite number of reciprocal lattice vectors, we can numerically solve for the eigenvalues and eigenfunctions with controllable accuracy.
To see this, we substitute the expansions into Eq. \eqref{eq:mod}:
\begin{equation}
    \mathcal{H}_b u_{n,\mathbf{k}}(\mathbf{x}) = \left[-\frac{1}{2}(\nabla + i\mathbf{k})^2 + \sum_{\mathbf{K'} \in \Lambda^*} V_{\mathbf{K'}} e^{i\mathbf{K'}\cdot\mathbf{x}}\right] \sum_{\mathbf{K} \in \Lambda^*} c_{n,\mathbf{k}}(\mathbf{K}) e^{i\mathbf{K}\cdot\mathbf{x}}.
\end{equation}
Applying the operator and collecting terms, we have:
\begin{equation}
    \begin{aligned}
        \mathcal{H}_b u_{n,\mathbf{k}}(\mathbf{x}) &= \sum_{\mathbf{K} \in \Lambda^*} c_{n,\mathbf{k}}(\mathbf{K}) \left[\frac{1}{2}|\mathbf{k} + \mathbf{K}|^2 e^{i\mathbf{K}\cdot\mathbf{x}}\right] + \sum_{\mathbf{K'} \in \Lambda^*} V_{\mathbf{K'}} e^{i\mathbf{K'}\cdot\mathbf{x}} \sum_{\mathbf{K} \in \Lambda^*} c_{n,\mathbf{k}}(\mathbf{K}) e^{i\mathbf{K}\cdot\mathbf{x}} \\
        &= \sum_{\mathbf{K} \in \Lambda^*} c_{n,\mathbf{k}}(\mathbf{K}) \left[\frac{1}{2}|\mathbf{k} + \mathbf{K}|^2 e^{i\mathbf{K}\cdot\mathbf{x}}\right] + \sum_{\mathbf{K},\mathbf{K'} \in \Lambda^*} V_{\mathbf{K'}} c_{n,\mathbf{k}}(\mathbf{K}) e^{i(\mathbf{K} + \mathbf{K'})\cdot\mathbf{x}}.
    \end{aligned}
\end{equation}
Projecting onto the basis function $e^{i\mathbf{Q}\cdot\mathbf{x}}$ for some $\mathbf{Q} \in \Lambda^*$ and integrating over the unit cell $\Omega$, we obtain:
\begin{equation}
    \begin{aligned}
        \int_{\Omega} e^{-i\mathbf{Q}\cdot\mathbf{x}} \mathcal{H}_b u_{n,\mathbf{k}}(\mathbf{x}) d\mathbf{x} &= \sum_{\mathbf{K} \in \Lambda^*} c_{n,\mathbf{k}}(\mathbf{K}) \left[\frac{1}{2}|\mathbf{k} + \mathbf{K}|^2 \int_{\Omega} e^{i(\mathbf{K} - \mathbf{Q})\cdot\mathbf{x}} d\mathbf{x}\right] \\
        &+ \sum_{\mathbf{K},\mathbf{K'} \in \Lambda^*} V_{\mathbf{K'}} c_{n,\mathbf{k}}(\mathbf{K}) \int_{\Omega} e^{i(\mathbf{K} + \mathbf{K'} - \mathbf{Q})\cdot\mathbf{x}} d\mathbf{x}.
    \end{aligned}
\end{equation}
Using the orthogonality of the plane-wave basis, we find that the integrals evaluate to $|\Omega|$ when the exponents vanish and zero otherwise. This leads to the following matrix equation for the coefficients:
\begin{equation}
    \frac{1}{2}|\mathbf{k} + \mathbf{Q}|^2 c_{n,\mathbf{k}}(\mathbf{Q}) + \sum_{\mathbf{K'} \in \Lambda^*} V_{\mathbf{K'}} c_{n,\mathbf{k}}(\mathbf{Q} - \mathbf{K'}) = E_{n,\mathbf{k}} c_{n,\mathbf{k}}(\mathbf{Q}).
\end{equation}
By truncating the set of reciprocal lattice vectors $\mathbf{K}$ to a finite subset, we can numerically solve this matrix eigenvalue problem to obtain approximations of the eigenvalues $E_{n,\mathbf{k}}$ and the coefficients $c_{n,\mathbf{k}}(\mathbf{K})$, from which we can reconstruct the Bloch functions.

\begin{figure}[h!]
    \centering
    \includegraphics[width=1\textwidth]{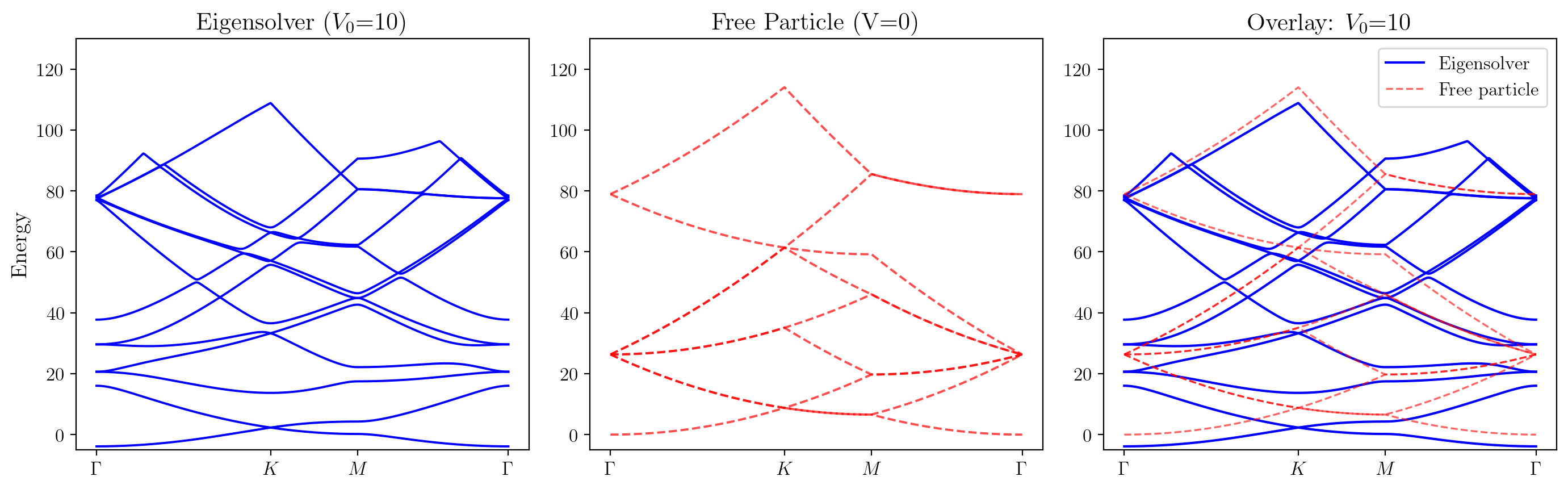}
    \caption{A side by side comparison of the band structure due to a strong honeycomb potential (left) and the free electron structure (right). The stronger potential induces significant band gaps and clearly show to the characteristic Dirac point at the K points, while the free electron model exhibits parabolic bands without gaps. Dispersion obtained via plane-wave expansion method and analytic solution respectively.}
    \label{fig:fig1}
\end{figure}
\subsection{Physics-Informed Neural Networks for Eigenvalue Problems}

Physics-informed neural networks (PINNs), first introduced by Raissi et al. \cite{raissi_physics_2017}, leverage the universal approximation capabilities of neural networks \cite{pinkus_approximation_1999} to solve forward and inverse PDE problems through direct incorporation of governing physical relations into the optimization objective. This is typically implemented by defining a neural network that approximates the solution to the PDE and then formulating a multi-term weighted objective, known as the loss function, which penalizes deviations in the form of PDE residuals, boundary conditions, initial conditions (if applicable), and data mismatch (if available) as \textit{soft constraints} on the network's output. 

By training the network to minimize this loss through iterated sampling of the input space and reweighting of the parameters, typically with a stochastic optimizer such as Adam, the model learns to satisfy the underlying physics of the problem. These methods have the key advantage of being mesh-free, unlike traditional numerical methods such as finite element or finite difference methods, as well as the promise of greater efficiency in high-dimensional settings, which suffer from the so-called \textit{curse of dimensionality} \cite{han2018solving, raissi_physics_2017}.

This approach has since been extended and applied to a variety of domains with notable success, including fluid dynamics \cite{cai2021physics}, material science \cite{faroughi2024physics}, and more recently, quantum mechanics \cite{jin2020unsupervised,jin_physics-informed_2022,hsu_equation-driven_nodate}. In recent years, there has been increasing attention to physics-informed learning for eigenvalue problems, such as are frequently encountered in physics and engineering, exemplified here by the time-independent Schrödinger equation. In this setting, the goal is to learn eigenvalues and eigenfunctions of differential operators or to find eigenfunctions for a predetermined eigenvalue. Notably, Jin et al. developed a PINN framework for solving quantum eigenvalue problems with square-well and Coulomb potentials, as well as a scanning neural network approach to learn multiple energetic eigenstates of the infinite square well and quantum harmonic oscillator \cite{jin2020unsupervised,jin_physics-informed_2022}. 

However, to the best of our knowledge, there has been only limited exploration of physics-informed neural solvers for more general two-dimensional periodic quantum eigenproblems, as most works have tended to focus on one-dimensional wells or quantum harmonic oscillators. The setting of honeycomb lattice potentials and their spectral properties provide a rich testbed for studying the interplay between symmetry and band structure degeneracy in quantum materials. The closest such work is that of Hsu et al. \cite{hsu_equation-driven_nodate}, who studied equation-driven (i.e. unsupervised) neural solvers for the Bloch eigenvalue problem for square lattices using constant and pseudo-atomic potentials. In a similar spirit, this thesis aims to extend this line of work by developing a physics-informed neural framework specifically tailored to honeycomb lattice potentials and their associated dispersion features, with an exploration of the benefits of transfer learning with pretraining on nearly-free electron potentials to generalize to strongly varying potentials.
\section{Description of Tasks}
\subsection{Learning Eigenpairs for Nearly-Free Honeycomb Potentials}
Using an \textit{ab initio} physics-informed learning framework, we aim to recover the band structure and complex Bloch functions for weak honeycomb lattice potentials simultaneously, which can be treated as small perturbations of the free-particle Hamiltonian. This involves training neural networks to approximate the Bloch eigenfunctions and eigenvalues across the Brillouin zone, enforcing the Schrödinger equation, Bloch periodicity, and normalization constraints through a composite loss function.
\subsection{Generalizing to Strong Potentials through Transfer Learning}
Building upon the models trained on nearly-free electron potentials, we explore transfer learning techniques to adapt these networks to handle strongly varying honeycomb lattice potentials, with possible extensions to pseudo-atomic potentials that more closely resemble real materials. This involves fine-tuning the pre-trained networks on new potential landscapes, say, by tuning $V_0$ in a parameterized potential form such as $V(\mathbf{x}) = V_0 \sum_{i=1}^{3} \cos(\mathbf{b_i}\cdot\mathbf{x})$, where $\mathbf{b_i}$ are the reciprocal lattice vectors, and examining the model's ability to capture changes in band structure topology, including the persistence or modification of Dirac points and band gaps.
\subsection{Benchmarking with Plane Wave Expansion}
To validate the accuracy of the physics-informed neural solvers, we implement a well-converged classical plane-wave expansion method to compute the band structure and Bloch functions for the same honeycomb lattice potentials as explained in Section 2.3. The results from the neural networks will be compared against these benchmarks for band structure recovery over a high-symmetry path in the Brillouin zone, as well as comparison of learned Bloch functions at selected $\mathbf{k}$-points.
\section{Methods}
\subsection{Neural Parameterization of Bloch Eigenfunctions}
We begin by defining a neural network to approximate the Bloch functions $u_{n,\mathbf{k}}(\mathbf{x})$ over the unit cell $\Omega$ for a given quasi-momentum $\mathbf{k}$ in the Brillouin zone. The model takes as input the spatial coordinates $\mathbf{x} \in \Omega$ and the quasi-momentum $\mathbf{k}$ in reciprocal space, and outputs the complex-valued Bloch function $u_{n,\mathbf{k}}(\mathbf{x})$. To handle the complex amplitude, we represent the output as two real-valued components corresponding to the real and imaginary parts:
\begin{equation}
    \mathcal{N}_1(\mathbf{x}, \mathbf{k}; \theta) = \begin{pmatrix} u_{n,\mathbf{k}}^{\text{Re}}(\mathbf{x}) \\ u_{n,\mathbf{k}}^{\text{Im}}(\mathbf{x}) \end{pmatrix},
\end{equation}
as well as an associated eigenvalue network:
\begin{equation}
    \mathcal{N}_2(\mathbf{k}; \phi) = E_{n,\mathbf{k}},
\end{equation}
where $\theta$ and $\phi$ denote the trainable parameters of the respective networks. The full Bloch wavefunction can then be reconstructed using Eq. \eqref{eq:bloch}:
\begin{equation}
    \psi_{n,\mathbf{k}}(\mathbf{x}) = e^{i\mathbf{k}\cdot\mathbf{x}} \left[u_{n,\mathbf{k}}^{\text{Re}}(\mathbf{x}) + i u_{n,\mathbf{k}}^{\text{Im}}(\mathbf{x})\right].
\end{equation}
We note a subtlety here relating to the band index $n$, which can be handled in two ways, principally: either by training separate networks for each band of interest using an iterated approach with orthogonality constraints in the spirit of Jin et al. \cite{jin_physics-informed_2022}, or by employing a single network with single band-output, where the higher energy bands and Bloch modes can be discovered simultaneously through extended sampling of the Brillouin zone, as in Hsu et al. \cite{hsu_equation-driven_nodate}. In this work, we primarily focus on the latter approach, as it allows for more efficient training and exploration of the band structure topology with minimal architectural complexity. During training, these networks are optimized jointly to minimize a composite loss function that enforces the governing Schrödinger equation, normalization constraints, and Bloch periodicity conditions and are coupled through the residuals of the PDE.
\subsection{Loss Formulation}
The training objective is formulated as a weighted sum of multiple loss components that enforce the physical constraints of the Floquet-Bloch eigenvalue problem. Specifically, we define the total loss $\mathcal{L}$ as:
\begin{equation}
    \mathcal{L} = \mathcal{L}_{\text{PDE}} + \lambda_{\text{norm}} \mathcal{L}_{\text{norm}} + \lambda_{\text{BC}} \mathcal{L}_{\text{BC}},
\end{equation}
where $\mathcal{L}_{\text{PDE}}$ penalizes the residual of the Schrödinger equation, $\mathcal{L}_{\text{norm}}$ enforces normalization of the Bloch functions, and $\mathcal{L}_{\text{BC}}$ imposes the Bloch periodicity conditions. The hyperparameters $\lambda_{\text{norm}}$ and $\lambda_{\text{BC}}$ control the relative importance of these terms, and this ``weighting'' of the loss function is of great importance to the success of training, and can be chosen according to the application, and one of the key challenges in physics-informed learning is the proper balancing of these terms to ensure convergence to physically valid solutions in a notoriously ill-conditioned optimization landscape \cite{krishnapriyan2021characterizing}. 

There is no universal recipe for the choice of weights which guarantees convergence to good minima in all settings, and ultimately they must be set empirically for each problem or chosen according to some adaptive self-weighting scheme, which typically relies on balancing the magnitudes of the gradients of each loss term or alternatively, examining the spectrum of the Neural Tangent Kernel (NTK), which governs the training dynamics of neural networks \cite{wang2023expertsguidetrainingphysicsinformed, jacot2018neural}. Note also that this loss defines an unsupervised learning framework, as it does not require any labeled data for supervision, relying solely on the governing physics of the problem, making it particularly well-suited for problems where data is scarce or expensive to obtain. The network itself produces the predictions of the Bloch modes and eigenvalues, which are then used to train the model by minimizing the total loss $\mathcal{L}$. Here, they have been set empirically as $\lambda_{\text{norm}} = 100$ and $\lambda_{\text{BC}} = 10$ based on preliminary experiments to balance the contributions of each term during training.
\subsubsection{Schrödinger PDE Residual}
At each iteration of training, we sample a batch of collocation points $\{\mathbf{x}_i\}_{i=1}^{N_b}$ uniformly at random from the unit cell $\Omega$ and a batch of quasi-momenta $\{\mathbf{k}_j\}_{j=1}^{N_b}$ uniformly from the Irreducible Brillouin Zone (IBZ) and its lattice translates. Passing these through the networks, we compute the mean squared residual of the modified Schrödinger equation in Eq. \eqref{eq:mod}:
\begin{equation}
    \mathcal{L}_{\text{PDE}} = \frac{1}{N_b} \sum_{i=1}^{N_b} \left| \mathcal{H}_b \mathcal{N}_1(\mathbf{x}_i, \mathbf{k}_i; \theta) - \mathcal{N}_2(\mathbf{k}_i; \phi) \mathcal{N}_1(\mathbf{x}_i, \mathbf{k}_i; \theta) \right|^2,
\end{equation}
where the Hamiltonian operator $\mathcal{H}_b$ is applied using automatic differentiation supported by \texttt{PyTorch} to compute the Laplacian and gradient terms with respect to the spatial inputs $\mathbf{x}$. Note that the eigenvalue problem as posed admits a trivial zero solution, which is avoided by the inclusion of the normalization loss term described next.
\subsubsection{Normalization Constraints}
To ensure that the learned Bloch functions are physically valid eigenstates, we enforce the following normalization condition over the unit cell $\Omega$:
\begin{equation}
    \mathcal{L}_{\text{norm}} = \frac{1}{N_n} \sum_{i=1}^{N_n} \left( \int_{\Omega} \left| \mathcal{N}_1(\mathbf{x}, \mathbf{k}_i; \theta) \right|^2 d\mathbf{x} - 1 \right)^2,
\end{equation}
where $\{\mathbf{k}_i\}_{i=1}^{N_n}$ are sampled quasi-momenta from the Brillouin zone, and in practice, the integral is numerically approximated using a trapezoidal rule over a fixed grid of points in $\Omega$. By unit normalizing the Bloch modes, we ensure that the eigenfunctions correspond to physically meaningful states with the added benefit that the squared amplitude can be interpreted as a probability density.
\subsubsection{Bloch Periodicity Conditions}
Due to Bloch's theorem, the wavefunctions must satisfy the pseudo-periodic boundary conditions in Eq. \eqref{eq:pseudo}. Or equivalently, the Bloch functions must be periodic over the unit cell. To enforce this, we sample points on the boundaries of the unit cell and compute the mean squared error between the values of the Bloch function at opposite boundaries:
\begin{equation}
    \mathcal{L}_{\text{BC}} = \frac{1}{N_b} \sum_{i=1}^{N_b} \left| \mathcal{N}_1(\mathbf{x}_i + \mathbf{a_j}, \mathbf{k}_i; \theta) - \mathcal{N}_1(\mathbf{x}_i, \mathbf{k}_i; \theta) \right|^2,
\end{equation}
\subsection{Sampling Strategy in Real and Reciprocal Space}
\begin{figure}[h]
    \centering
    \includegraphics[width=0.6\textwidth]{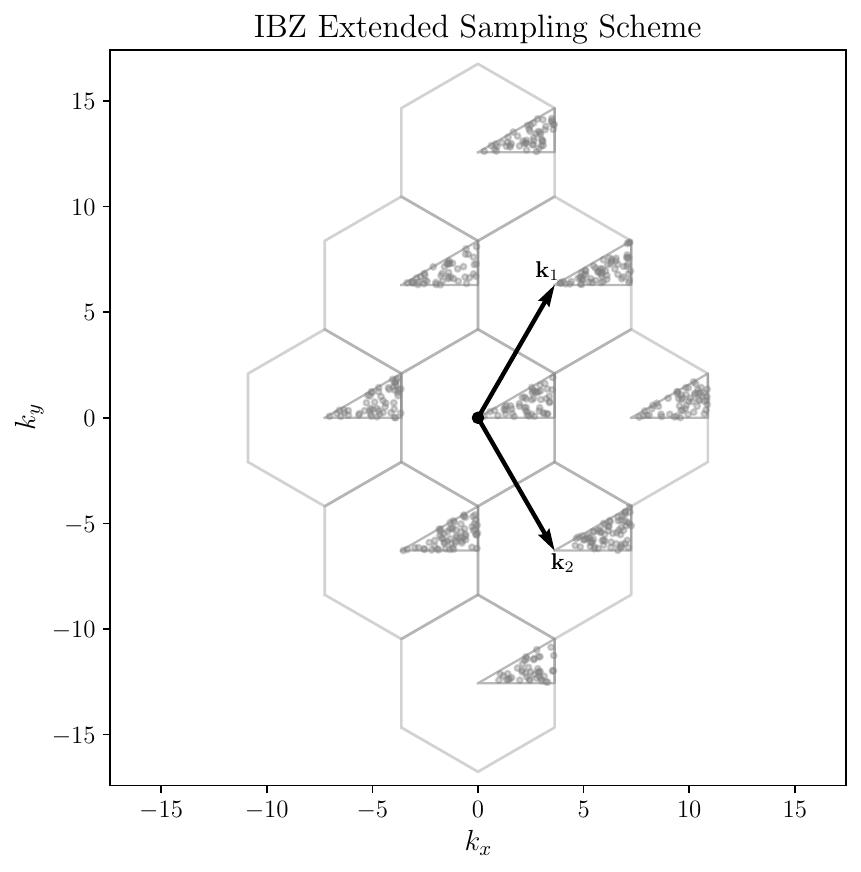}
    \caption{Sampling strategy in reciprocal space. Points are sampled uniformly within the Irreducible Brillouin Zone (IBZ) and its lattice translated copies to expose the network to higher quasi-momenta and energy bands.}
    \label{fig:fig1}
\end{figure}
In order to effectively train the neural networks, we employ a sampling strategy that covers both the spatial domain of the unit cell $\Omega$ and the quasi-momentum space of the Brillouin zone. For the spatial domain, we use uniform random sampling of collocation points within the unit cell spanned by the lattice vectors $\{\mathbf{a_1}, \mathbf{a_2}\}$. This ensures that the network learns to satisfy the Schrödinger equation and boundary conditions throughout the entire unit cell.

As for the quasi-momentum space, we focus on sampling points inside the Irreducible Brillouin Zone (IBZ), which is the smallest wedge of the Brillouin zone that, through symmetry operations, can generate the entire zone. This reduces the computational burden while still allowing the network to learn the essential features of the band structure. We sample $\mathbf{k}$-points uniformly within the IBZ and also from its lattice translated copies to expose the network to higher energy bands.
\subsection{Model Architecture}
While many architectural choices are possible for the networks $\mathcal{N}_1$ and $\mathcal{N}_2$, we opt for fully-connected feedforward neural networks (FNNs) due to their universal approximation capabilities and ease of implementation. The Bloch function network $\mathcal{N}_1$ consists of an input layer that takes the concatenated spatial coordinates $\mathbf{x}$ and quasi-momentum $\mathbf{k}$, followed by several hidden layers with smooth nonlinear activation functions (SiLU) and an output layer producing the real and imaginary components of the Bloch function. The eigenvalue network $\mathcal{N}_2$ is similarly structured but takes only the quasi-momentum $\mathbf{k}$ as input and outputs the corresponding eigenvalue $E_{n,\mathbf{k}}$. Dimensions of the hidden layer is set to 200 with 5 hidden layers for the Bloch net, and 64 with 3 hidden layers for the energy net. 
\subsection{Training and Optimization Details}
We train the networks using the Adam optimizer \cite{kingma2017adammethodstochasticoptimization} with an initial learning rate of $1e-3$, which is combined with a cosine annealing schedule to gradually reduce the learning rate over the course of training. The models are trained for a total of 20,000 epochs with a batch size of 3,000 collocation points sampled per epoch. At each epoch, we compute the total loss $\mathcal{L}$ and backpropagate the gradients to update the network parameters $\theta$ and $\phi$. To monitor convergence, we track the individual loss components as well as the overall loss throughout training. All experiments were conducted on a single NVIDIA RTX A6000 GPU with 48GB of memory.
\section{Analysis}
\subsection{Band Structure Recovery for Weak Honeycomb Potential}
\begin{figure}[h]
    \centering
    \includegraphics[width=0.6\textwidth]{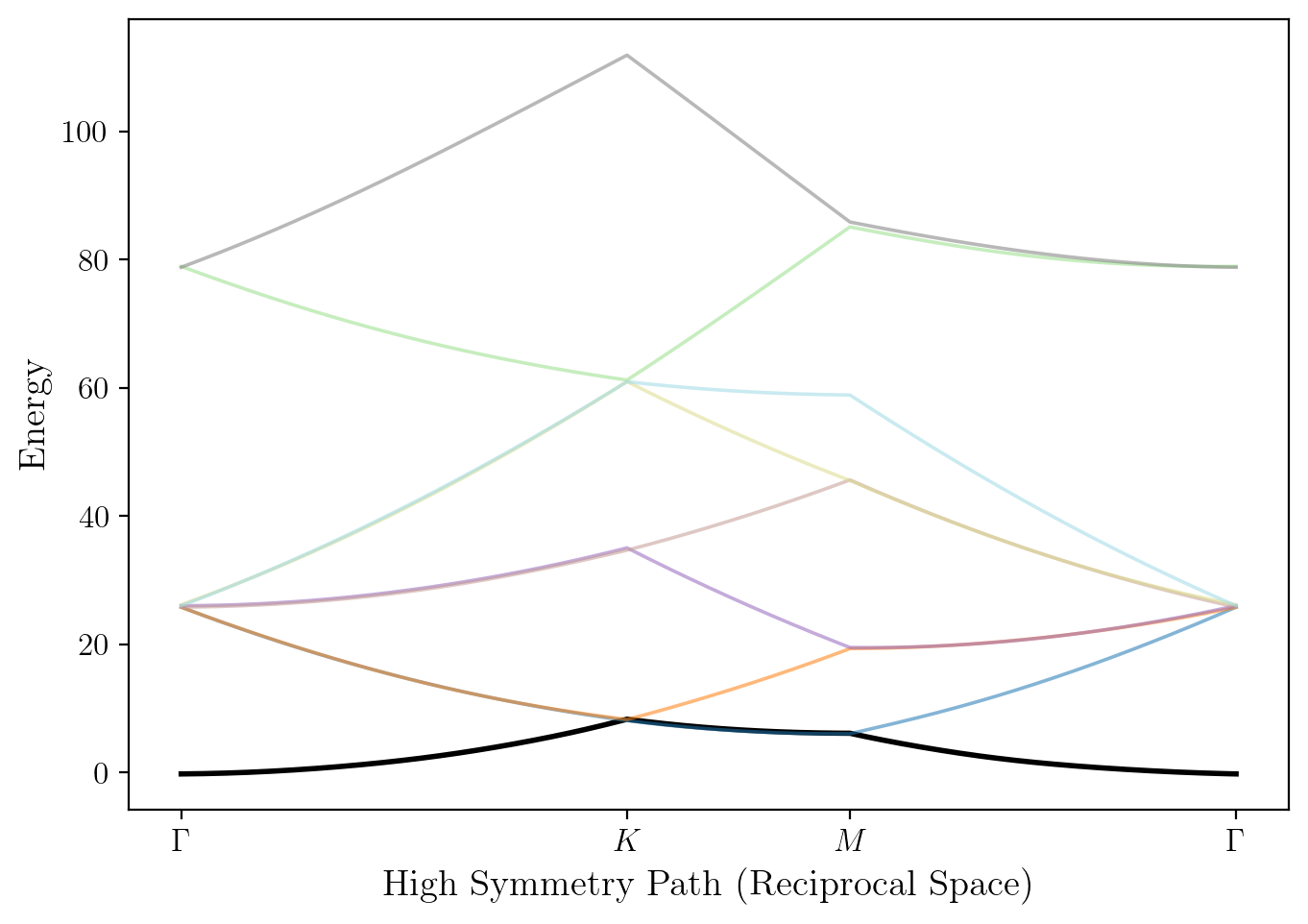}
    \caption{Band structure learned by the physics-informed neural network plotted along the high-symmetry path $\Gamma \rightarrow K \rightarrow M \rightarrow \Gamma$ in the Brillouin zone for a weak honeycomb potential with $V_0 = 1$. The learned dispersion curves match closely with those obtained from the plane-wave expansion method, and closely approximate the free particle dispersion due to the weak potential.}
    \label{fig:fig1}
\end{figure}
After training the networks on a weak honeycomb lattice potential of the form $V(\mathbf{x}) = V_0 \sum_{i=1}^{3} \cos(\mathbf{b_i}\cdot\mathbf{x})$ with $V_0 = 1$, we evaluate the learned band structure by running inference on the trained eigenvalue network $\mathcal{N}_2(\mathbf{k}; \phi)$ over a dense sampling of $\mathbf{k}$-points along the high-symmetry path $\Gamma \rightarrow K \rightarrow M \rightarrow \Gamma$ in the Brillouin zone and its extended zone copies. The resulting dispersion curves are then compared to those obtained from the plane-wave expansion method to assess the accuracy of the neural solver in capturing the band structure topology.
\subsection{Comparison with Plane-Wave Expansion}
\begin{figure}[h!]
    \centering
    \includegraphics[width=0.6\textwidth]{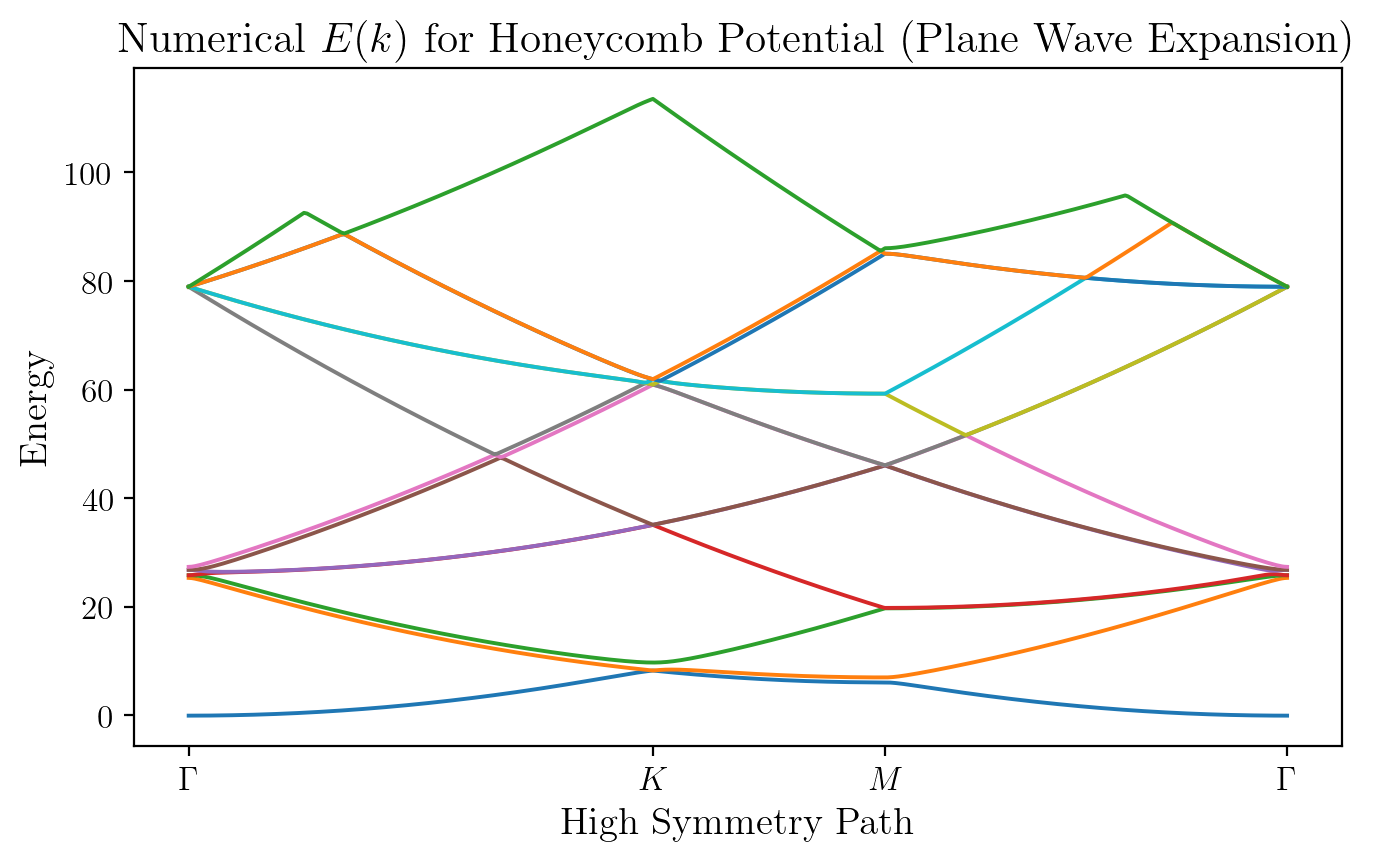}
    \caption{Numerically computed band structure using the plane-wave expansion method for the same weak honeycomb potential with $V_0 = 1$. The results serve as a benchmark for evaluating the accuracy of the physics-informed neural network's predictions.}
    \label{fig:fig1}
\end{figure}
The band structure obtained from the physics-informed neural network shows excellent agreement with the benchmark results from the plane-wave expansion method, as illustrated in Figures 5 and 6. The learned dispersion curves closely follow the expected behavior for a nearly-free electron model, with minor deviations attributable to the weak potential perturbation.
\subsection{Error Metrics and Convergence Analysis}
\begin{figure}[h]
    \centering
    \includegraphics[width=0.7\textwidth]{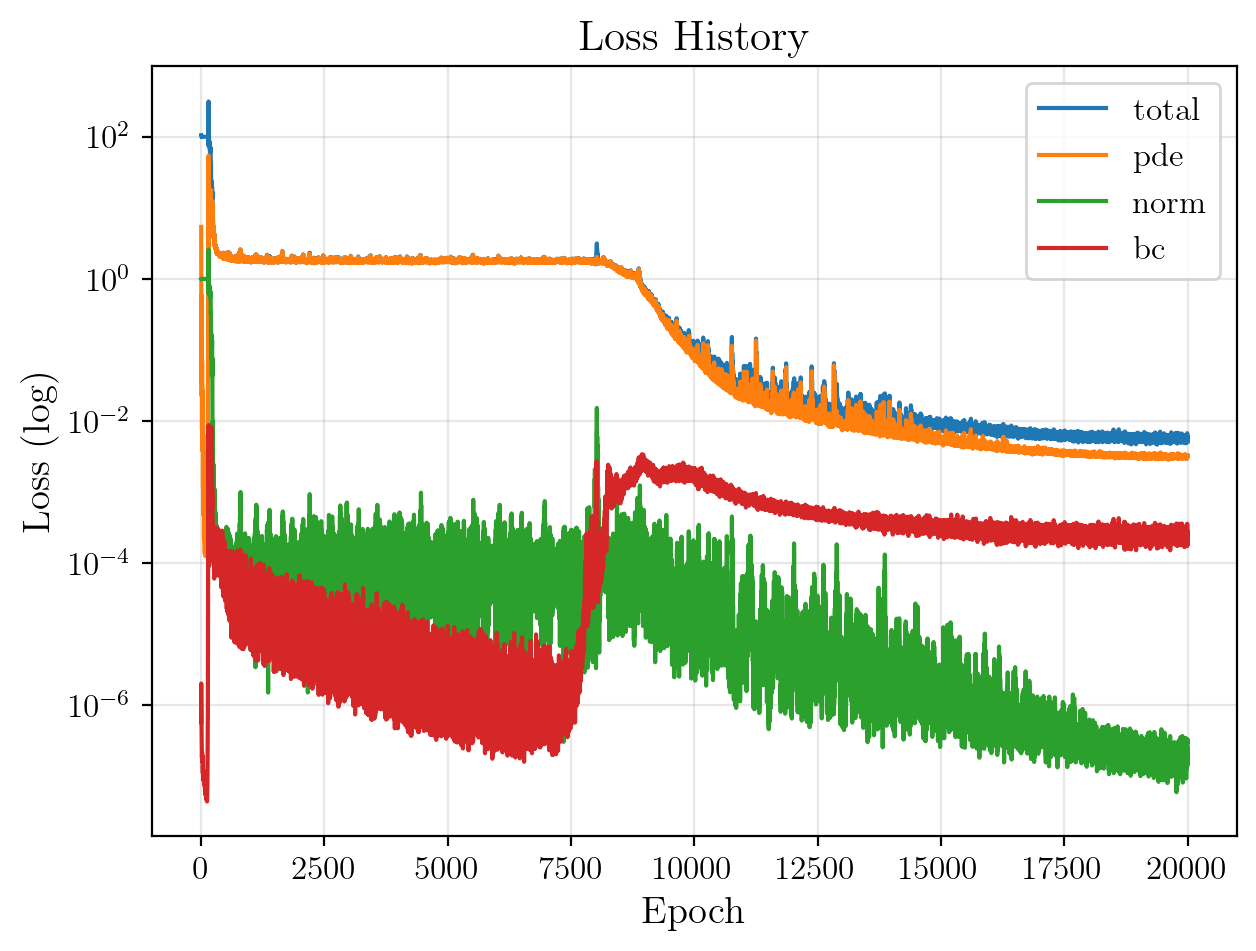}
    \caption{Training loss curves for the physics-informed neural network on the weak honeycomb potential. The total loss and individual components are plotted over training epochs, demonstrating convergence behavior. Around epoch 7500, the network begins to converge to a stable solution and prioritizes minimizing the PDE residual temporarily before settling into a balanced minimization of all loss components.}
    \label{fig:fig1}
\end{figure}
Loss metrics are tracked during training to assess convergence behavior both for the overall loss and individual components, which is crucial for diagnosing potential issues in optimization. The final values for the PDE residual, normalization error, and boundary condition error are reported in Table \ref{tab:loss_values}, demonstrating the model's ability to satisfy the governing physics of the problem.
\begin{table}[h]
    \centering
    \begin{tabular}{|c|c|}
        \hline
        Loss Component & Final Value \\
        \hline
        Total Loss ($\mathcal{L}$) & $5.2 \times 10^{-3}$ \\
        PDE Residual ($\mathcal{L}_{\text{PDE}}$) & $3.3 \times 10^{-3}$ \\
        Normalization Error ($\mathcal{L}_{\text{norm}}$) & $3.0 \times 10^{-7}$ \\
        Boundary Condition Error ($\mathcal{L}_{\text{BC}}$) & $1.9 \times 10^{-4}$ \\
        \hline
    \end{tabular}
    \caption{Final loss values after training on weak ($V_0 = 1$) honeycomb potential.}
    \label{tab:loss_values}
\end{table}
\subsection{Visualization of Bloch Functions}

To further validate the learned Bloch functions, we visualize their spatial profiles at selected $\mathbf{k}$-points, such as the high-symmetry points $\Gamma$, $K$, and $M$. The squared amplitude $|u_{n,\mathbf{k}}(\mathbf{x})|^2$ is plotted over the unit cell to illustrate the probability density distribution of the electrons in these states. We include one representative state produced by the model which shows good qualitative agreement with the corresponding Bloch modes computed via the plane-wave expansion method, as shown in Figure 8.

\begin{figure}[h]
    \centering
    \includegraphics[width=0.6\textwidth]{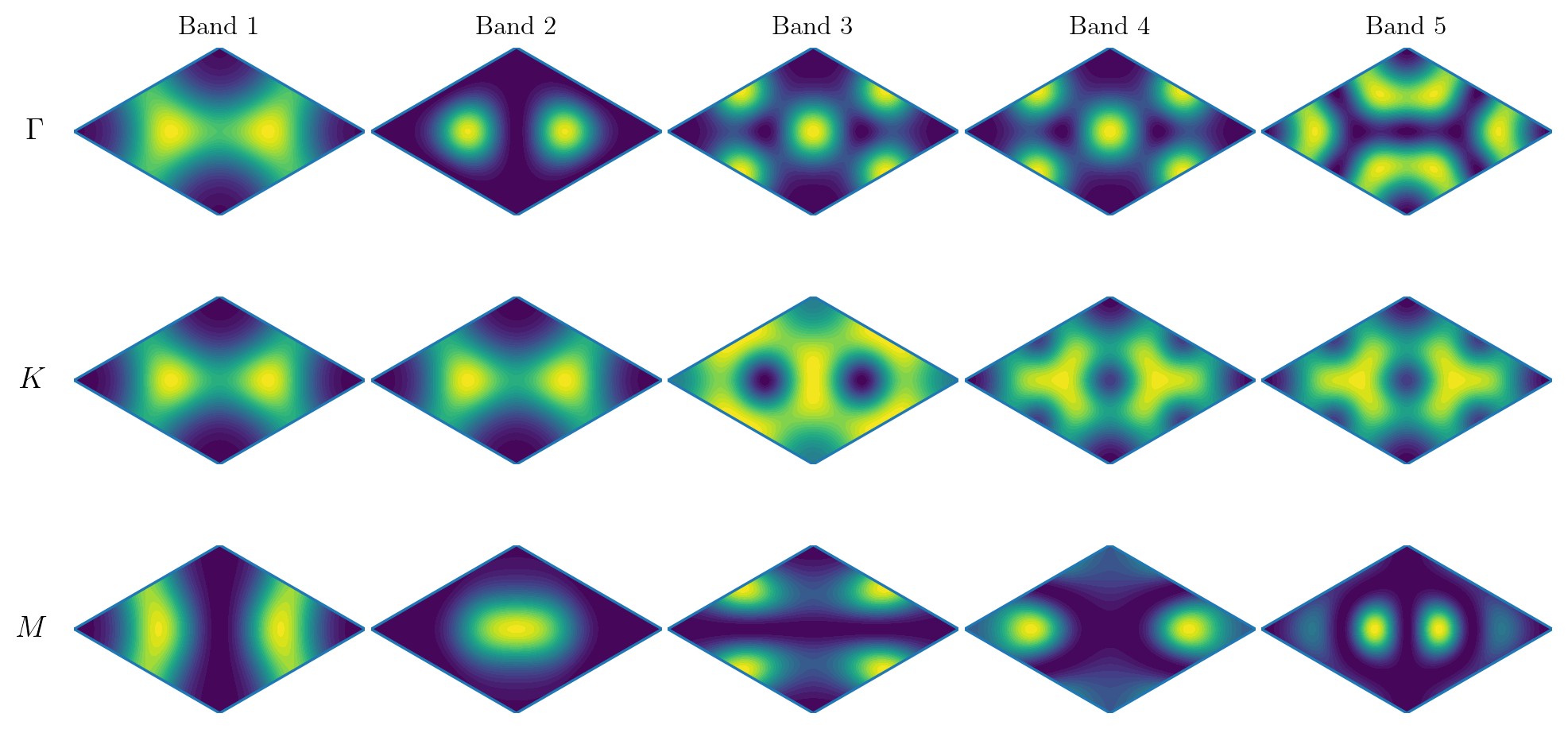}
    \hfill
    \includegraphics[width=0.25\textwidth]{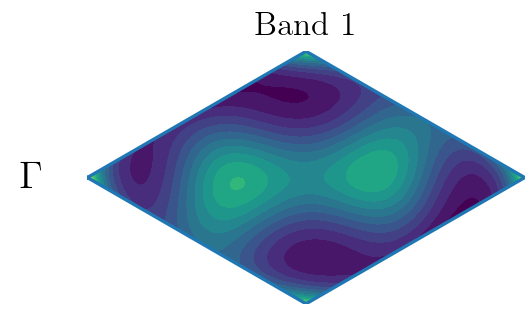}
    \caption{Representative Bloch modes plotted as $|u_{n,\mathbf{k}}(\mathbf{x})|^2$ over the unit cell at the high-symmetry points computed using the plane-wave expansion, as well as a representative Bloch state learned by the model at the $\Gamma$ point after fine-tuning on $V_0=10$ (right).}
    \label{fig:fig1}
\end{figure}
\section{Concluding Remarks}
To conclude, we have developed a physics-informed neural framework for solving the Floquet-Bloch eigenvalue problem for honeycomb lattice potentials, demonstrating its ability to accurately recover band structures and Bloch functions for weak potentials. The model's predictions closely match benchmark results from the plane-wave expansion method, validating the effectiveness of the approach. 

Future work will explore extensions to stronger potentials including pseudo-atomic well-like potentials and perhaps explicit handling of the band index through orthogonality constraints. Fourier feature embeddings, as in \cite{sallam2023use}, and other architectural enhancements may also be investigated to improve convergence and accuracy. Potentially, L-BFGS could be tried after convergence with Adam to improve the precision on the approach here. Overall, this work highlights the potential of physics-informed neural networks as a powerful tool for studying quantum materials with complex periodic potentials. 
\section{Acknowledgements}
We would like to thank \textbf{Professor Michael Weinstein} for the useful discussions that helped us to improve the quality and direction of this thesis, for his foundational contributions to the analysis of honeycomb lattice potentials and corresponding Dirac cones, without which this thesis would not have been possible, and for his continuous support and mentorship throughout my time at Columbia.

We would also like to thank \textbf{Professor Kathleen McKeown} for her support and assistance with computational resources used during this project, for her ongoing guidance, and for the formative impact she has had on my development as a researcher during my time at Columbia.

Finally, we thank \textbf{Professor Kui Ren} for his insightful comments and suggestions that enhanced the clarity of this work.

\newpage
\bibliography{MSThesis.bib}
\bibliographystyle{siam}

\end{document}